\documentclass{article}

\usepackage{arxiv}

\usepackage[utf8]{inputenc} 
\usepackage[T1]{fontenc}    
\usepackage{hyperref}       
\usepackage{url}            
\usepackage{booktabs}       
\usepackage{amsfonts}       
\usepackage{nicefrac}       
\usepackage{microtype}      
\usepackage{lipsum}
\usepackage{graphicx}
\graphicspath{ {./images/} }

\usepackage{graphicx}

\usepackage{subcaption}
\usepackage{cleveref}   

\title{Feature Reuse and Fusion for Real-time Semantic segmentation}

\author{
 Sixiang Tan \\
  College of Information Science and Engineering, Xinjiang University \\
}

\begin{document}
\maketitle
\begin{abstract}
For real-time semantic segmentation, how to increase the speed while maintaining high resolution is a problem that has been discussed and solved. 
Backbone design and fusion design have always been two essential parts of real-time semantic segmentation. We hope to design a light-weight network based on previous design experience and reach the level of state-of-the-art real-time semantic segmentation without any pre-training.
To achieve this goal, a encoder-decoder architectures are proposed to solve this problem by applying a decoder network onto a backbone model designed for real-time segmentation tasks and designed three different ways to fuse semantics and detailed information in the aggregation phase.
We have conducted extensive experiments on two semantic segmentation benchmarks. Experiments on the Cityscapes and CamVid datasets show that the proposed FRFNet strikes a balance between speed calculation and accuracy. It achieves 72.0\% Mean Intersection over Union (mIoU) on the Cityscapes test dataset with the speed of 144 FPS on a single RTX 1080Ti card.The Code is available at https://github.com/favoMJ/FRFNet

\end{abstract}


\section{Introduction}
Semantic segmentation, also known as scene parsing, predicts dense labels for all pixels in an image, as the core issue in remote sensing mapping, autonomous driving, medical image analysis and other application fields. Such real-world applications not only demand competitive performance but also have a strict requirement for low latency. Therefore, designing a semantic segmentation network with efficient inference speed and high accuracy becomes a challenging issue. 
To achieve high accuracy, Semantic segmentation models can usually pre-training to improve generalization. DDRNet[1] all the modules have been pre-trained on ImageNet, BiSenet[2], DDPNet[3], etc. prove that better results can be obtained after pre-training. We hope that without the introduction of external resources, the model can still achieve real-time and high-precision results. Because the real-time segmentation model is slight, many training data usually achieve better results, unlike other model training methods. We use the traversal data set for training, and experiments prove that this method has achieved good results on the test data set. 
We investigate how to generate multilevel representations with negligible computational overhead. In mainstream semantic segmentation, fusion with attention is usually inevitable, but we think layers of different depths have different semantic information. After fusing enough semantic information, we should use simple feature extraction methods to fuse features to accelerate the speed. In our decoding framework, we have made some improvements to the current attention. Three different methods are proposed for the fusion of semantics and details to achieve real-time segmentation.
The backbones of previous excellent networks have powerful function extraction functions, such as ResNet[4] ,  ResNeXt[5] ,  DenseNet[6] , Res2Net[7]. To achieve real-time speed and adapt to segmentation tasks, some works like CenterMask[8], DDPNet[3], DDRNet[1] designed a dedicated semantic segmentation network as the backbone network, and these backbones usually use the reuse of features to improve efficiency.
Based on the above observation, we propose a novel network architecture specially designated for real-time semantic segmentation, which is presented in Figure 1. 
We develop a Multi-Frequency Attention Module (MFAM) to capture long-range relations efficiently. To strike a better balance between accuracy and efficiency, we follow the principle of simplicity when designing models. We propose a light-weight but powerful backbone and a simple and practical framework to achieve fast and accurate semantic segmentation. 

Our main contributions are summarized as follows:
We design a light-weight and robust backbone with the proposed Dense-Reuse module(DRM) to sufficiently aggregate multi-scale contexts.
We propose three different ways to fuse semantics and detailed information in the aggregation phase.
We achieve impressive results on the benchmarks of Cityscapes\cite{cordts2016cityscapes} and CamVid\cite{brostow2008segmentation} benchmarks without any pre-trained model. It can run on high-resolution images (512× 1024) at 144 FPS on a single RTX 1080Ti card and obtain the results of 72.0\% mIoU on the Cityscapes test dataset.

\begin{figure}[htbp]
	\centering

	\begin{minipage}[t]{0.8\textwidth}
	\includegraphics[width=1\textwidth]{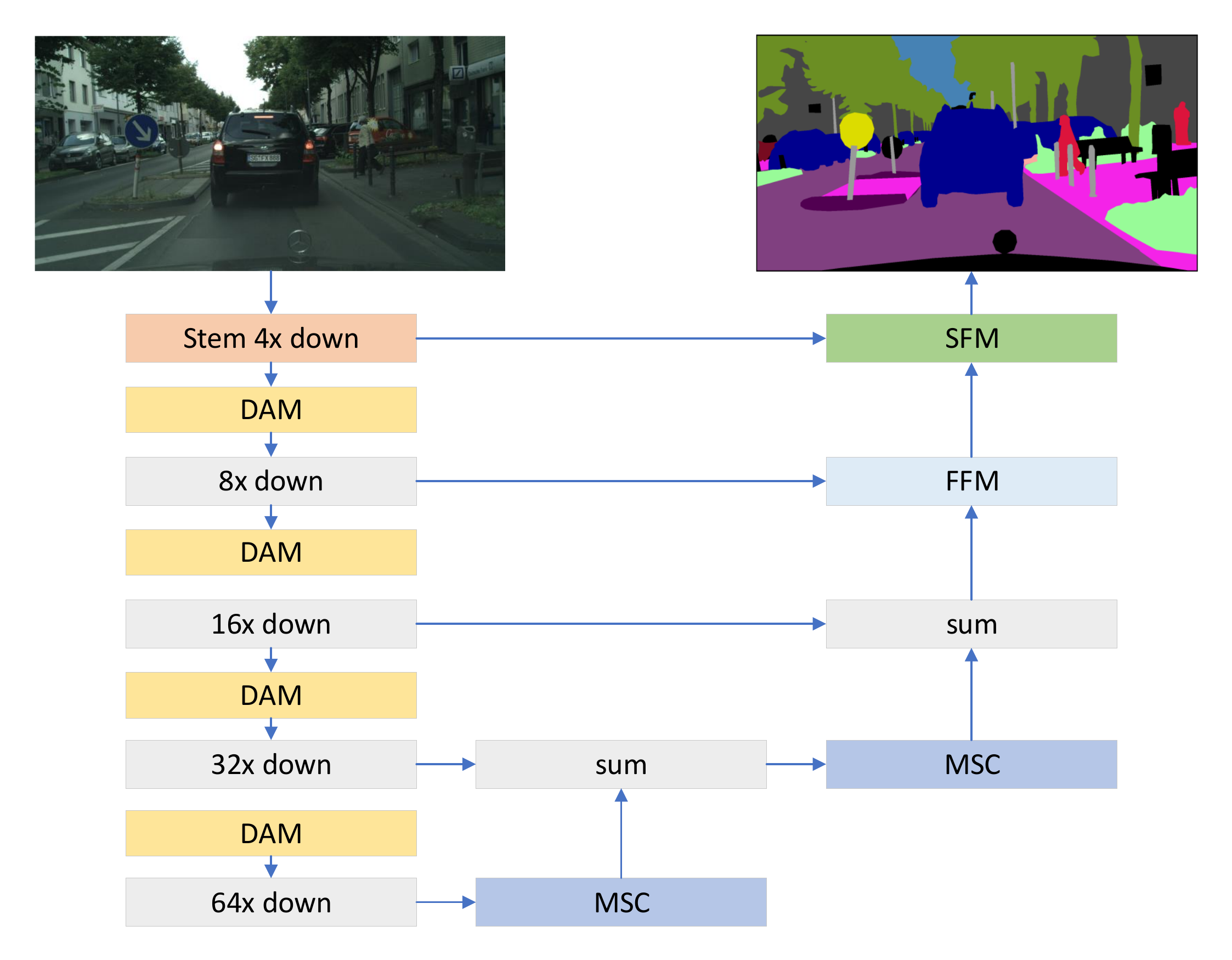}
\end{minipage}%
	

	\centering
	\caption{An overview of the Feature Reuse and Fusion for real-time semantic segmentation}
\end{figure}

\section{Relate Work}
\label{sec:headings}
Recently, many new methods have been used in semantic segmentation. In this section, we introduce some methods related to our work. i.e., real-time semantic segmentation method, feature fusion, feature enhancement and self-attention model.

\subsection{Real-time semantic segmentation}
The main purpose of real-time segmentation is to obtain high-quality predictions while obtaining low-latency inference speed. ENet[9] presents a network with down-samplings to pursue the ultimate rate. BiSeNet[2] decouple the function of spatial information preservation and receptive field offering into two paths. ERFNet[10]  focuses on better accuracy with a deeper network that uses factorized convolutions and residual connections to remain efficient. The light-weight model saves storage space, memory and calculations. ESPNetv2[11] use the EESP Unit to pursuit a larger receptive field with fewer parameters. CGNet[12] use CG block to capture contextual information in all stages.  
Our model follows the light-weight style to achieve real-time segmentation.

\subsection{Self-attention Model. }
The self-attention model can capture long-term dependencies and has been used by many models. DANet[13] use the self-attention mechanism to units spatial and channel information separately. ACFNet[13] propose a self-attention model to make different pixels adaptively focus on the different class centre. However, these methods need to generate substantial attention maps, which adds much computational overhead. To reduce the complexity of the self-attention mechanism, AttaNet[13] introduce a self-attention model to capture long-range dependencies. CCNet[13] leverages two crisscross attention modules to generate sparse connections for each position. BiSeNet[2] propose a specific attention refinement module to refine the features of each stage. Fcanet[14] extends channel attention to the frequency domain.
To capture long-term relationships more effectively, we also extract channel features in different frequency domains. Unlike Fcanet, which uses simple blocks to obtain frequency domain information, our method uses fine-grained methods to get more information.

\subsection{Feature fusion}
The traditional feature fusion method is usually sum up these features. 
BiSeNet[2] propose a feature fusion module to fuse detail and context information.
Attanet[15] use each pixel to choose individual contextual information from multi-level features in the aggregation phase. 
EfficientFCN [16] observes that the fusion of multi-scale feature maps usually leads to better performance.
We believe that there is enough semantic information after high-level fusion, so attention fusion is only used in the low-level, and simple fusion is used in the fusion stage where more details are needed.

\section{Method}
The overall network architecture of the proposed FRFNet is shown in Figure 1. As we can see, our FRFNet is a convolutional network using a single branch architecture. Two key modules are then introduced separately. To obtain more semantic information, we propose a more effective backbone network based on VoVNet2 that shows better performance and faster speed than ResNet. And we introduced three different methods to fuse semantics and detailed information at different stages.

\subsection{Backbone Design}
In this subsection, we mainly discuss the components used to build the backbone of FRFNet. In this work, we aim to design a backbone that gets the best compromise between accuracy and efficiency.

When designing real-time segmentation networks, many methods adopt depth-wise separable and factorized convolution. Instead of using depth-wise separable or factorized convolution, FRFNet is building with traditional convolution. 

\noindent\textbf{Transition Layer.} Transition layer is used to reduce the size of feature maps and compress the model. Like Inception-v2[17], we use two parallel stride blocks: P and C. P is a maximum pooling layer to avoid a representational bottleneck. 
FRFNet keeps the number of input channels and output channels the same in all transition layers to fully exploit dense connectivity. This design facilitates feature reuse throughout the whole network.

\begin{figure*}[ht!]
	\centering
	\begin{subfigure}{0.4\textwidth}
		\centering   
		\includegraphics[width=1\linewidth]{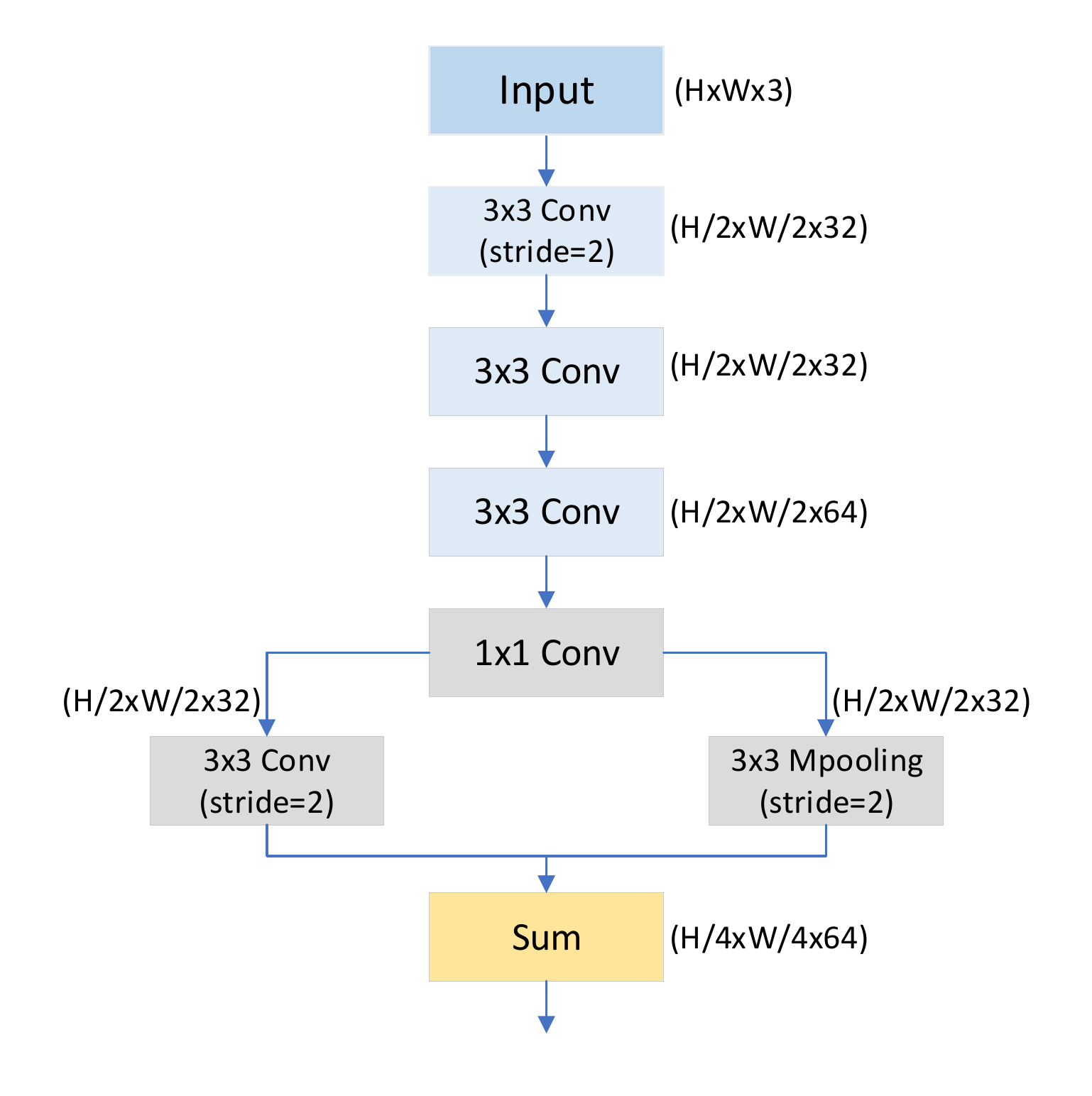}
		\caption{}
		\label{fig:sub1}
	\end{subfigure}   
	\begin{subfigure}{0.4\textwidth}
		\centering   
		\includegraphics[width=\linewidth]{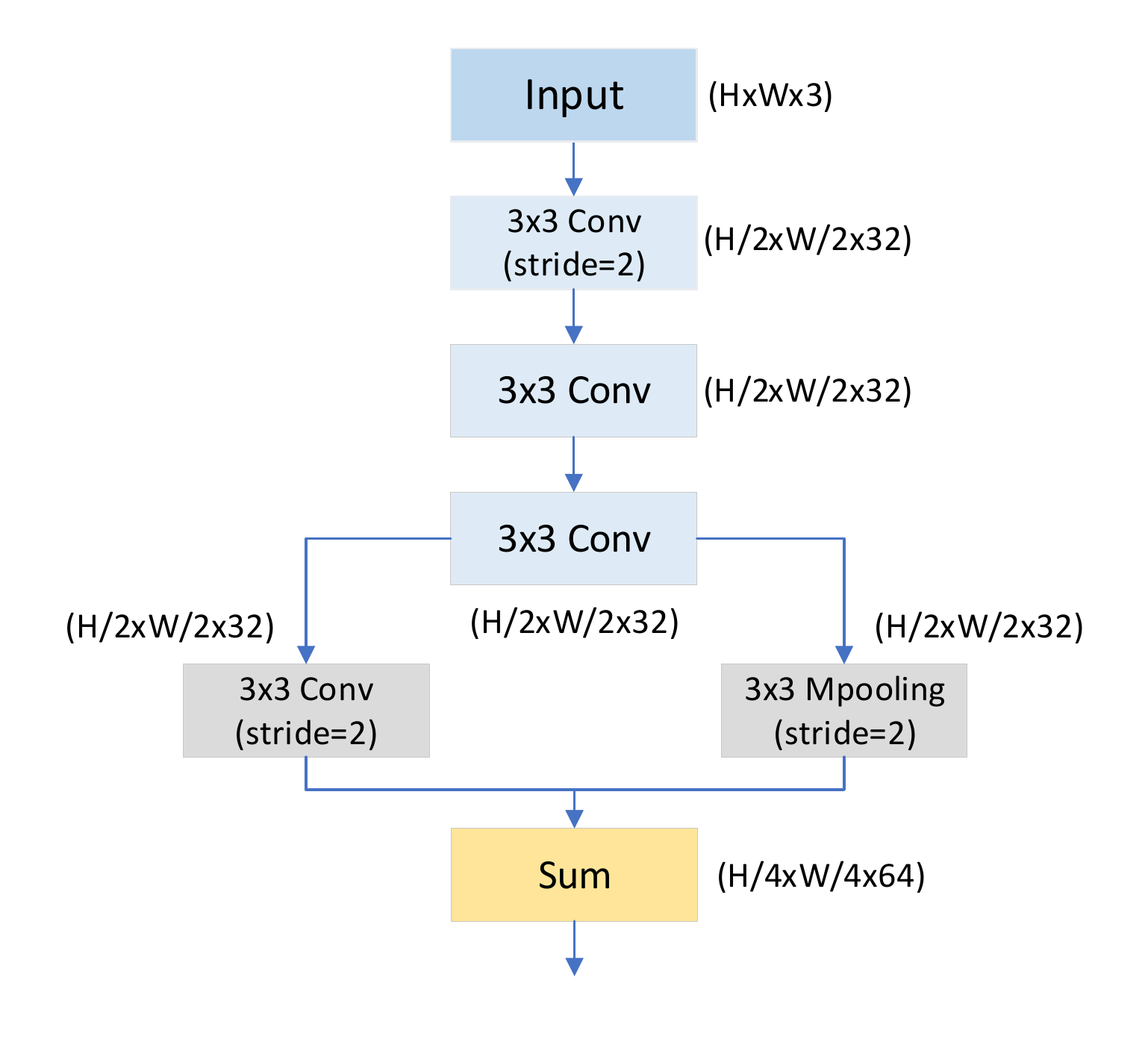}
		\caption{}
		\label{fig:sub2}
	\end{subfigure}
	\caption{
		\label{fig:total}
	Stem Block. From left to right are (a) FRFNet Stem Block.  (b) FRFNet-slim Stem Block.
	}
\end{figure*}

\noindent\textbf{Stem Block.}
The stem is used to reduce the input size, which usually involves multiple downsampling operations. Inspired by BiSeNet [2], we use stem blocks as the first stage of semantic branching,  as illustrated in Figure 2. It uses two different downsampling methods to reduce the feature representation. And the output functions of the two branches are connected in series as output. This structure has effective calculation cost and effective feature expression ability.

\begin{figure*}[ht!]
	\centering
	\begin{subfigure}{0.3\textwidth}
		\centering   
		\includegraphics[width=1\linewidth]{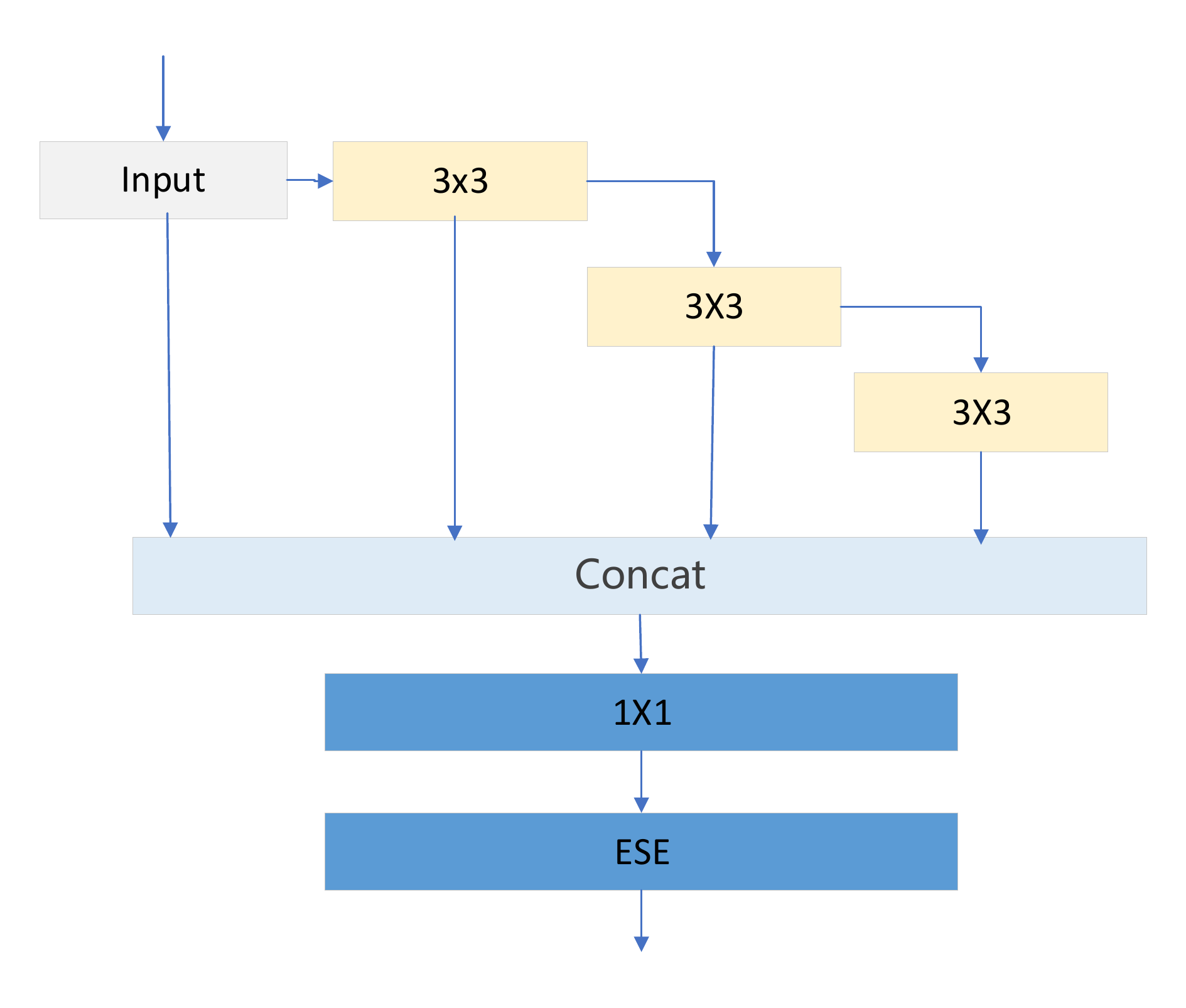}
		\caption{}
		\label{fig:sub1}
	\end{subfigure}   
	\begin{subfigure}{0.3\textwidth}
		\centering   
		\includegraphics[width=\linewidth]{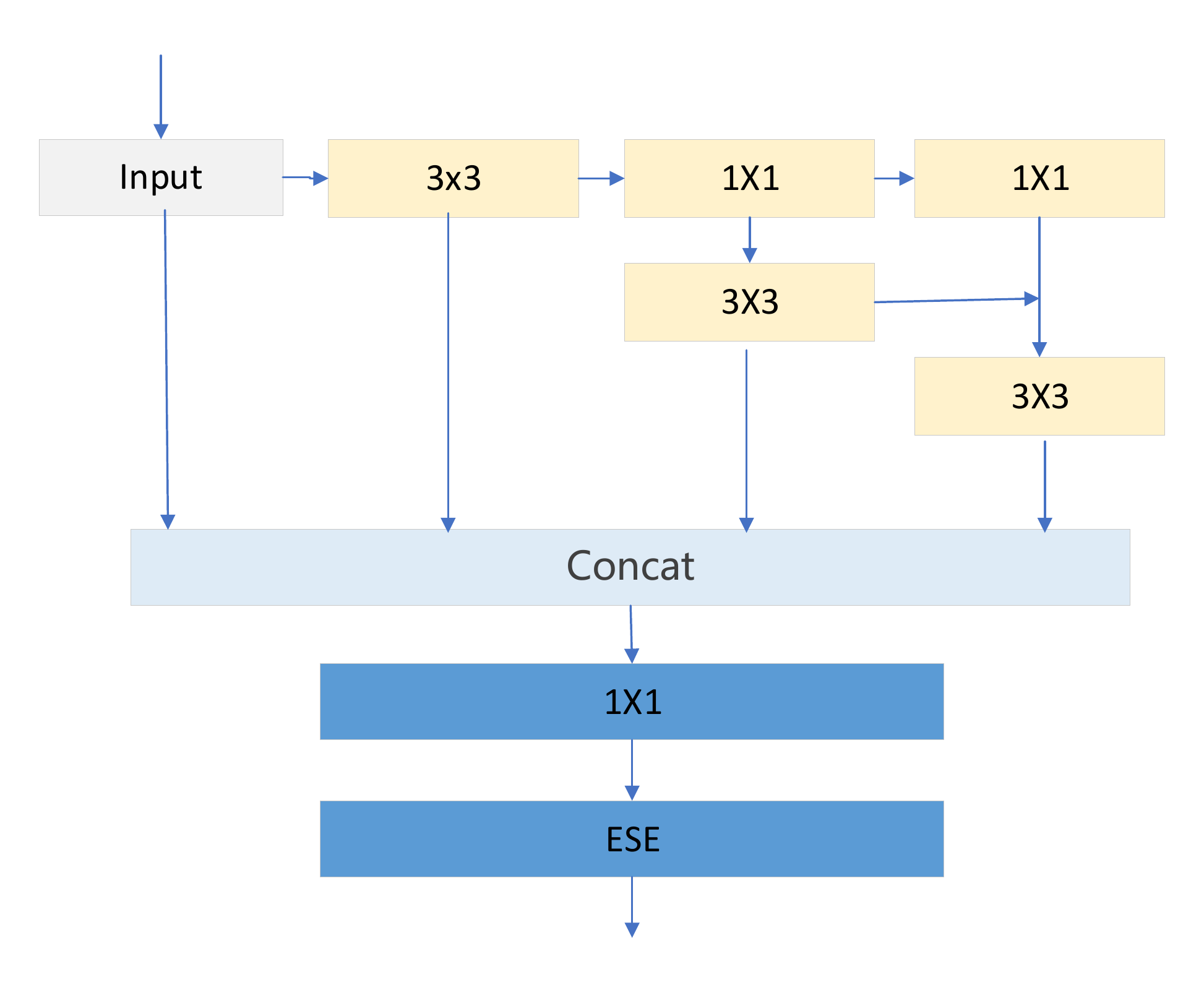}
		\caption{}
		\label{fig:sub2}
	\end{subfigure}
	\begin{subfigure}{0.3\textwidth}
		\centering   
		\includegraphics[width=\linewidth]{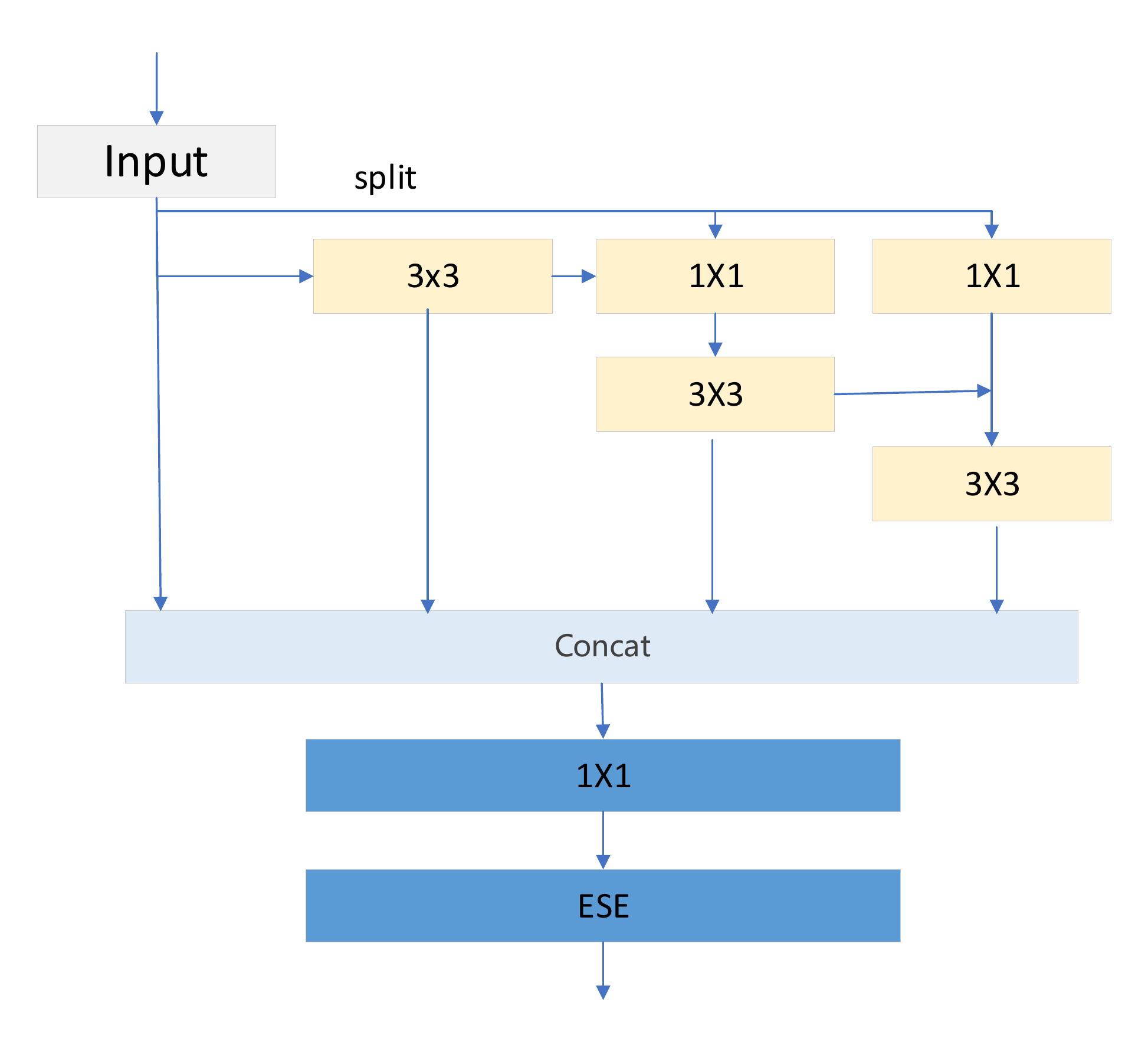}
		\caption{}
		\label{fig:sub2}
	\end{subfigure}
	\caption{
		\label{fig:total}
		An Comparison between our DAM module and OSA Module. From left to right are (a) OSA Module.  (b) DAM module. (c) DAM-Slim module.
	}
\end{figure*}

\noindent\textbf{Bottleneck.} To make full use of dense connections under a limited computing budget, 
OSA module[8] consists of consecutive convolution layers and aggregates the subsequent feature maps at once, which can capture diverse receptive fields, as shown in Figure 3(a).
Motivated from Res2Net[7],GhostNet[18], CenterMask[8]. We proposed a Dense Aggregation module (DAM), as depicted in Figure 3(b). The method of generating more feature maps from cheap operations is used, which is close to the representation ability of large and dense layers. And  We reused 3*3 convolutional feature layers to expand the network capacity to a certain extent without significantly increasing the complexity.

\subsection{Upsampling Module}
How to fuse the different information of the low-level and high-level predecessors have gone through a lot of excellent attempts. We believe that the integration of different levels requires different integration strategies. 
We think convolution on a large size will consume a lot of computation, which is not conducive to real-time segmentation and in our experience, training the detail information was difficult due to the domination of the strong semantic branch, so we focus on semantic extraction at the high layer, and detail extraction at the low layer.

\begin{figure*}[ht!]
	\centering
	\begin{subfigure}{0.3\textwidth}
		\centering   
		\includegraphics[width=1\linewidth]{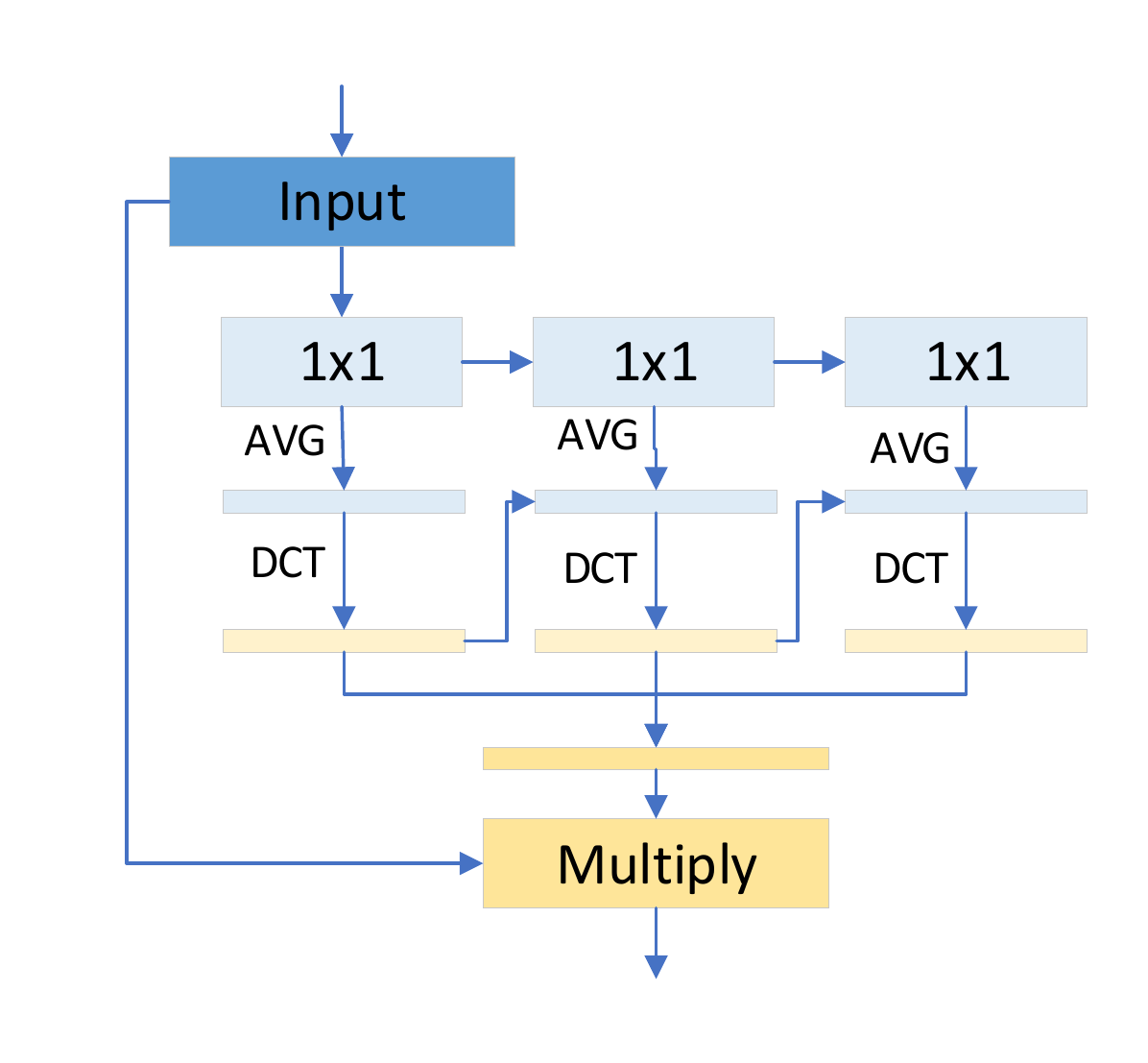}
		\caption{}
		\label{fig:sub1}
	\end{subfigure}   
	\begin{subfigure}{0.3\textwidth}
		\centering   
		\includegraphics[width=\linewidth]{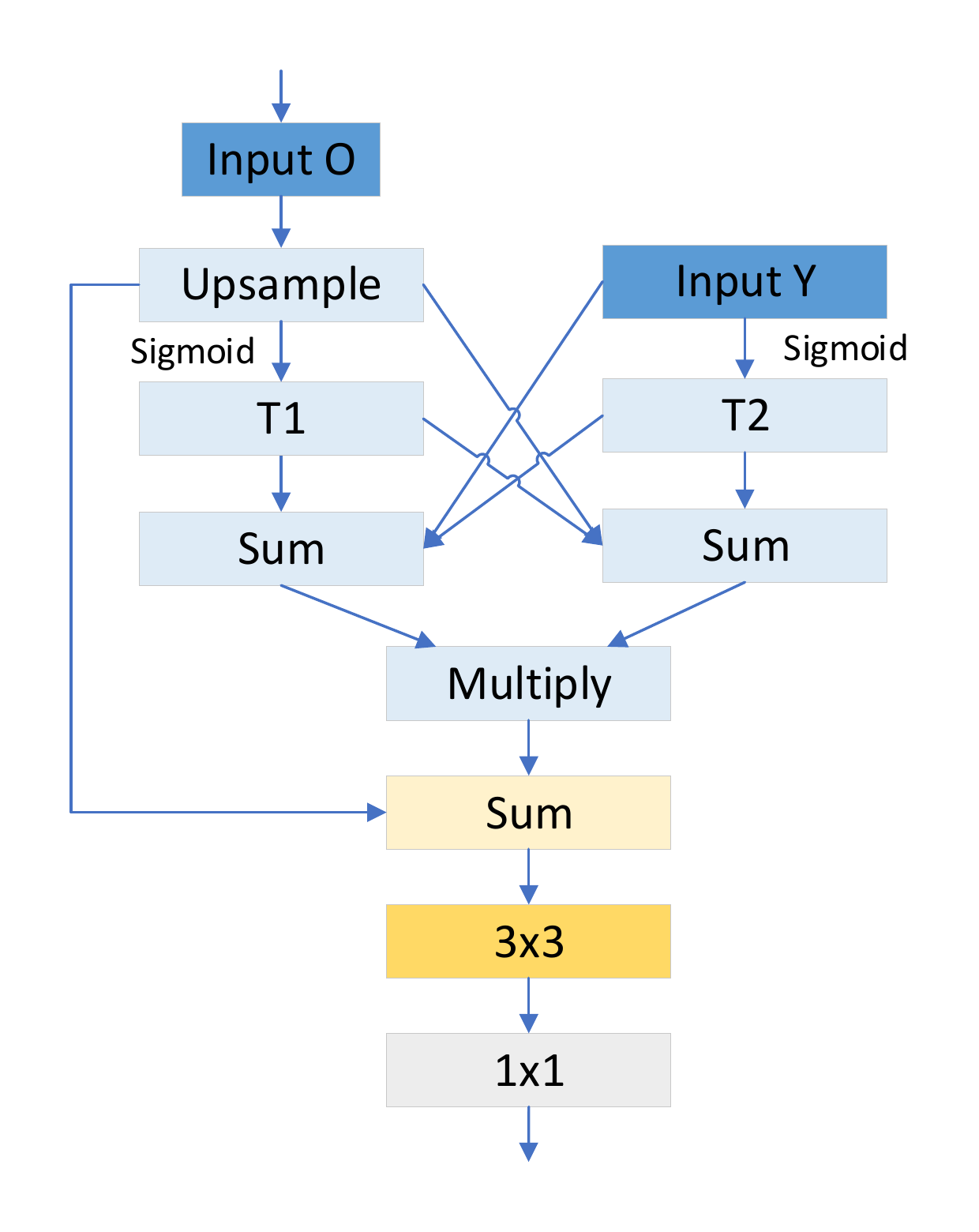}
		\caption{}
		\label{fig:sub2}
	\end{subfigure}
	\begin{subfigure}{0.3\textwidth}
		\centering   
		\includegraphics[width=\linewidth]{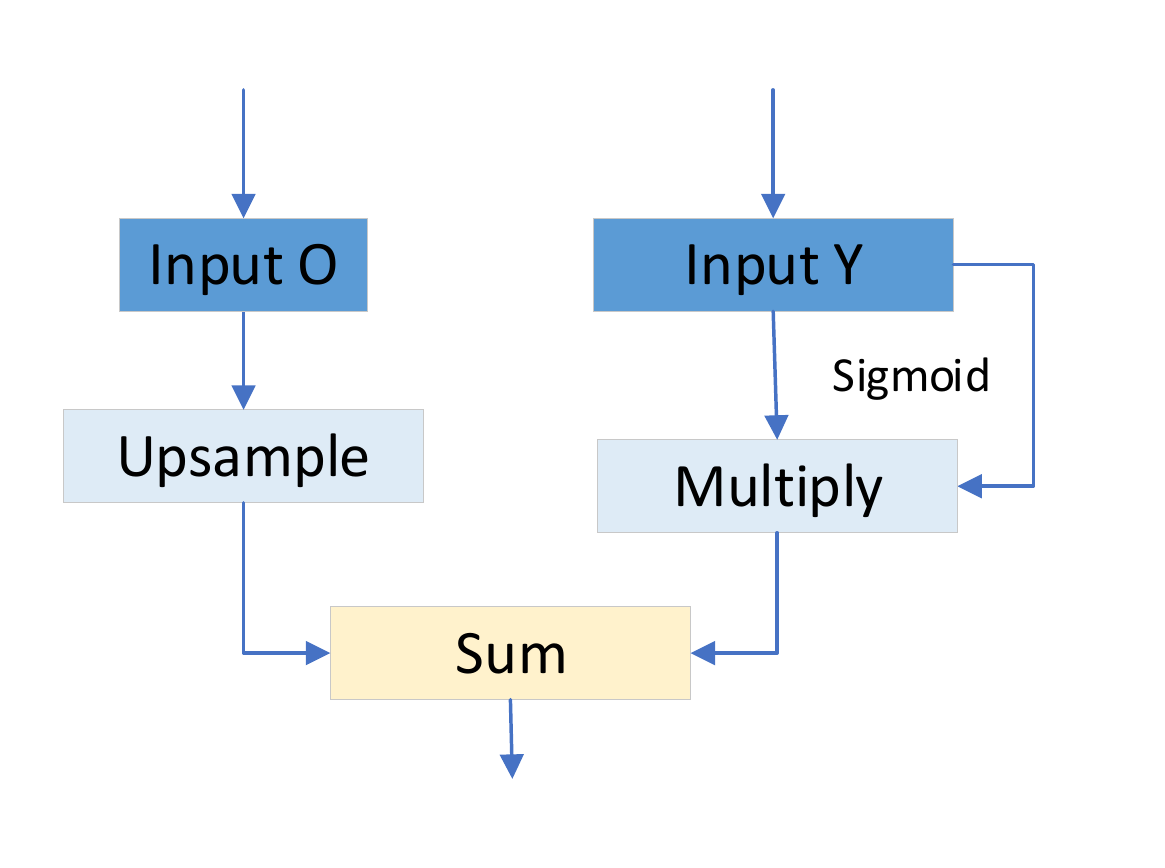}
		\caption{}
		\label{fig:sub2}
	\end{subfigure}
	\caption{
		\label{fig:total}
		Fusion Module. From left to right are (a) MSC Module (AVG means average pooling, DTC means special case of 2D).  (b) FFM Module. (c) SFM Module.
	}
\end{figure*}
Like many methods, we use attention enhancement to enhance the semantics in the last two feature layers. Previous work DANet[13], FACNet[14] has excellent attention modules. The attention model we use is shown in Figure 4(a). We refine the input into three small parts and use the Multi-Spectral Channel to obtain the attention.

In the second stage, we start to incorporate relevant details. The input to this module is two segmentation maps of size h×w×C: (1) the spatial feature map Y, and (2) the context feature map O. The output of the module is the detail-semantic fusion segmentation map Fig 4(b) depicts the fusion process, which includes the following steps. 

First, We obtain the probability feature map through sigmoid and get the probability map about semantics and details t1, t2 through the probability that the difference between the two is more significant than the threshold value. We set the threshold to 0.4 based on experimental experience. 
Next, the obtained feature maps are further fused through addition and 3x3 convolution, and finally, the final output feature map is obtained by 1x1 convolution.

In the third stage, as shown in Figure 4(c), we only use sigmoid and sum to obtain detailed information.

\section*{Experiments}
We evaluate our proposed network on the public datasets Cityscapes and Camvid. We first introduce the datasets and implementation protocol. Then, we conducted a series of ablation experiments on the Camvid validation dataset to prove the effectiveness of our network. Finally, we carry out comprehensive experiments on Cityscapes and CamVid benchmarks and compare them with other works, and we use the standard mIoU metric to report segmentation accuracy.

\subsection*{Datasets and Settings}

\noindent\textbf{Cityscapes.} The focus on semantic understanding of urban scenes. It has 5000 images of driving scenes in an urban environment (2975 train, 500 val, 1525 test). It has 19 categories of dense pixel annotation (97\% coverage), of which 8 have instance-level segmentation. The Cityscapes data set, the urban landscape data set, is a large-scale data set containing different stereo video sequences recorded in 50 different city street scenes.

\noindent\textbf{Camvid.} CamVid is the first video collection with semantic tags for the target category. It includes a total of 701 images, 367 for training, 101 for validation and 233 for testing. The image has a resolution of 360×480 and 11 semantic categories.

\subsection*{Implementation protocol}
All the experiments are performed with one 1080Ti GPU, CUDA 10.1  on the Pytorch platform. We follow the test code provided by SwiftNet for accurate measurement. 

The Adam optimizer is adopted to train our model. Expressly, the batch size is set to 16. We use cosine attenuation with an initial learning rate of $1e^{-4}$ and a minimum learning rate of $1e^{-6}$. We train the model for 1000 epochs on Cityscapes trainval dataset, and Camvid set 800 epochs on Camvid. For data augmentation, we employ random horizontal flip and mean subtraction. We randomly use the parameters between [0.5, 2] to transform the image to different scales. Then we randomly crop the resolution to 512×1024 on Cityscapes for training while the cropping resolution is 360×480 on Camvid.

\subsection*{Measure of Inference Speed}
The inference speed is measured on a single GTX 1080Ti GPU by setting the batch size to 1 and with CUDA 10.2, CUDNN 7.6 and PyTorch 1.4, and we report an average of 100 frames for the frames per second (FPs) measurement.

\subsection*{Ablation studies}
To further prove the effectiveness of the FRFNet, we conduct extensive ablation experiments on the validation set of Camvid and Cityscapes with different settings for FRFNet.

\begin{table}[!h]
	\setlength{\tabcolsep}{3mm}
	\caption{Accuracy and speed comparison of our method against other methods on Cityscapes test dataset. R18 stand for ResNet-18 with ImageNet pre-training.}
	\centering
	\begin{tabular}{@{}lccc@{}}
		\hline
		Model             & mIoU & Flops   & FPS   \\ 
		\hline
		SwiftNet-R18      & 75.5 & 103.37G  & 45.40 \\
		FRFNet-R18        & 70.7 & 13.72G   & 68.5  \\
		FRFNet-Danet-slim & 72.0 & 15.47G   & 144.4 \\
		FRFNet-Danet      & 75.1 & 25.40G   & 93.8  \\ 
		\hline
	\end{tabular}
\end{table}
\noindent\textbf{Backbone} 
Many previous real-time semantic segmentation architectures have used powerful feature extractors, such as ResNet-18. Good feature extraction has a profound impact on speed and accuracy. Therefore, we used ResNet-18 to compare with our backbone and studied the results (Table 1).
We can confirm that our backbone network is still lighter, faster, and more accurate from the table. 

\noindent\textbf{Feature Fusion Module}
\begin{table}[!h]
	\centering
	\caption{Ablation study results on camvid test set.}
	\label{tab:0}       
	\begin{tabular}{lc}
		\hline
		Network & mIoU(\%)  \\
		\hline
		Baseline & 65.3\\
		Baseline + MSC & 66.4 \\
		Baseline + MSC + FFM & 67.21  \\
		Baseline + MSC + FFM + SFM & 68.0  \\
		\hline
	\end{tabular}
\end{table}
It can be seen from Table 2 that an adequately designed fusion technology will have a significant impact on the final result. It also can be seen that the three fusion methods we developed have achieved a good balance between speed and accuracy. 
\begin{figure*}[ht]
	\centering 
	
	\begin{subfigure}[b]{0.3\textwidth}{
			\begin{minipage}[b]{1\linewidth} 
				\centering
				\includegraphics[width=\linewidth]{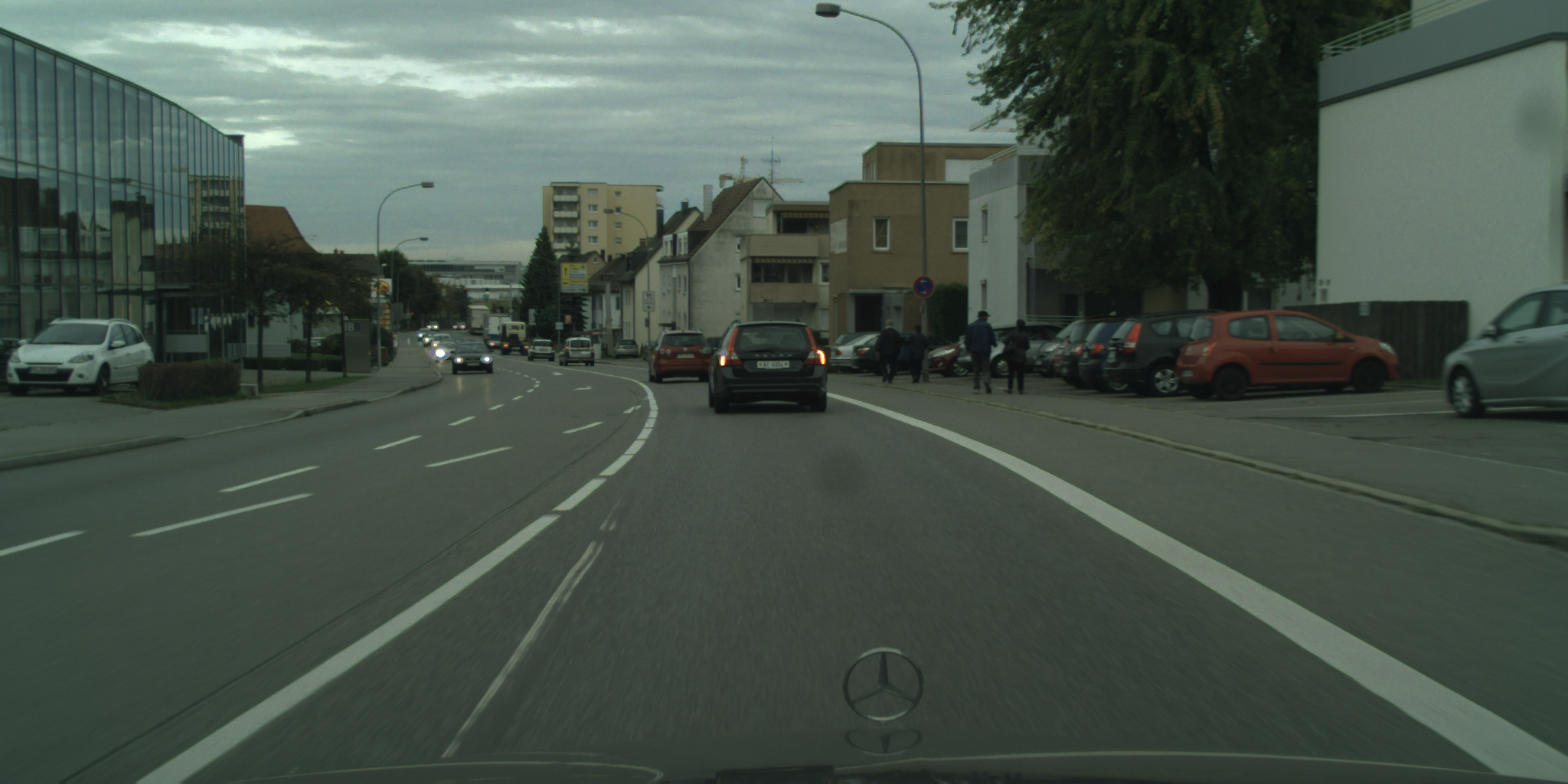}\vspace{8pt}
				\includegraphics[width=\linewidth]{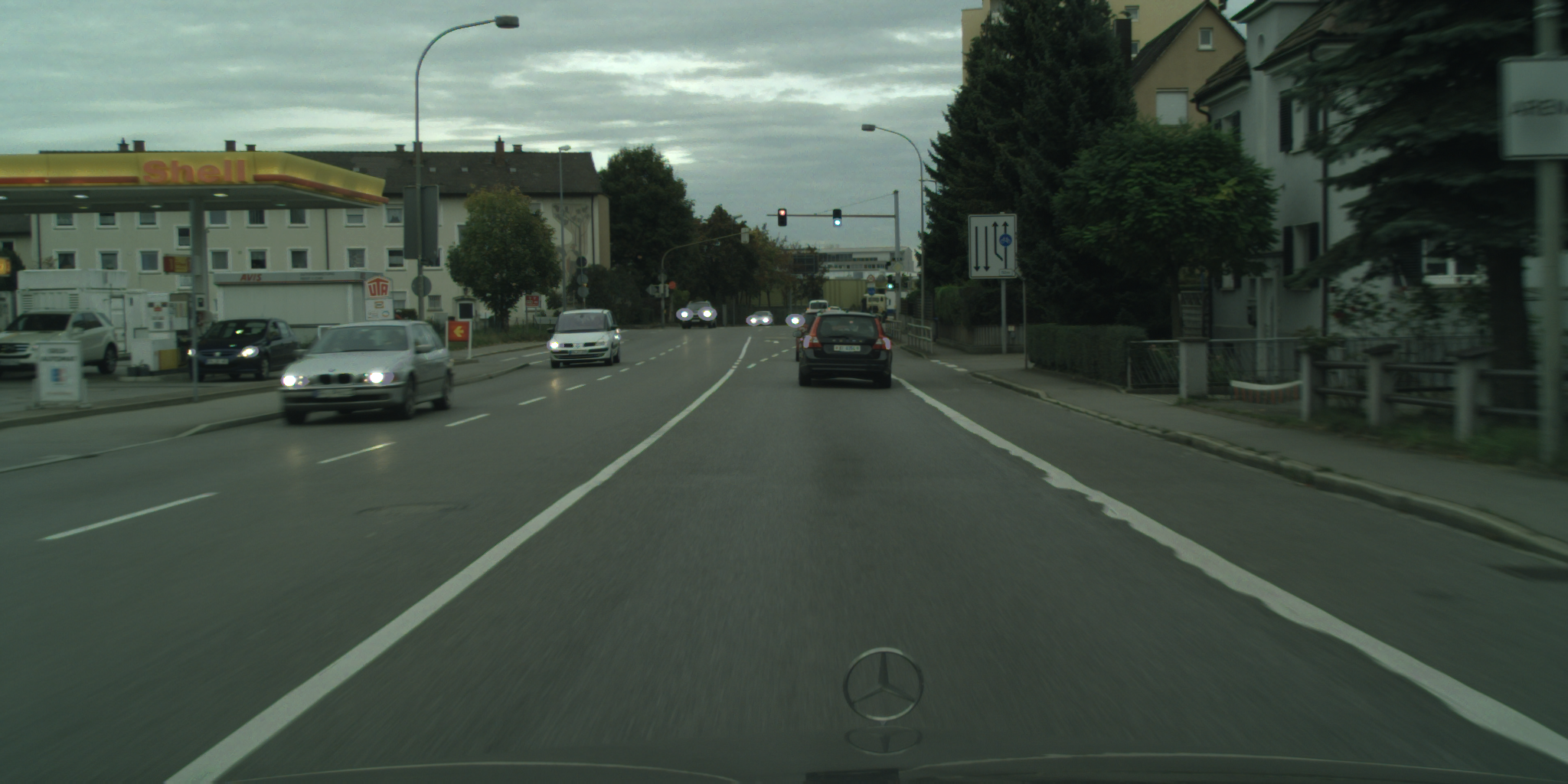}\vspace{8pt}
				\includegraphics[width=\linewidth]{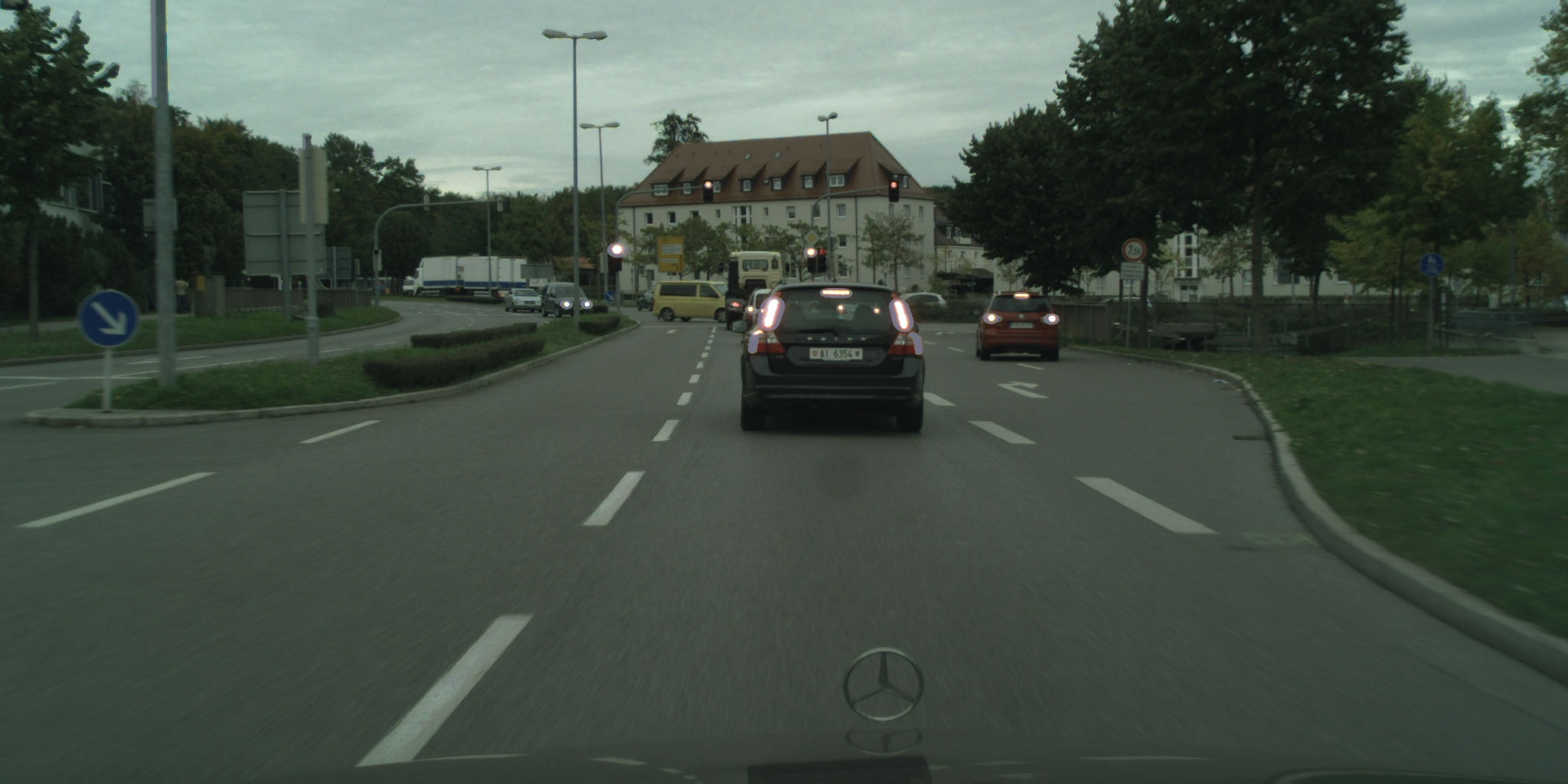}
				
			\end{minipage}
			\caption{Input image}
		}
	\end{subfigure} 
	\begin{subfigure}[b]{0.3\textwidth}{
			\begin{minipage}[b]{1\linewidth}
				\centering
				\includegraphics[width=\linewidth]{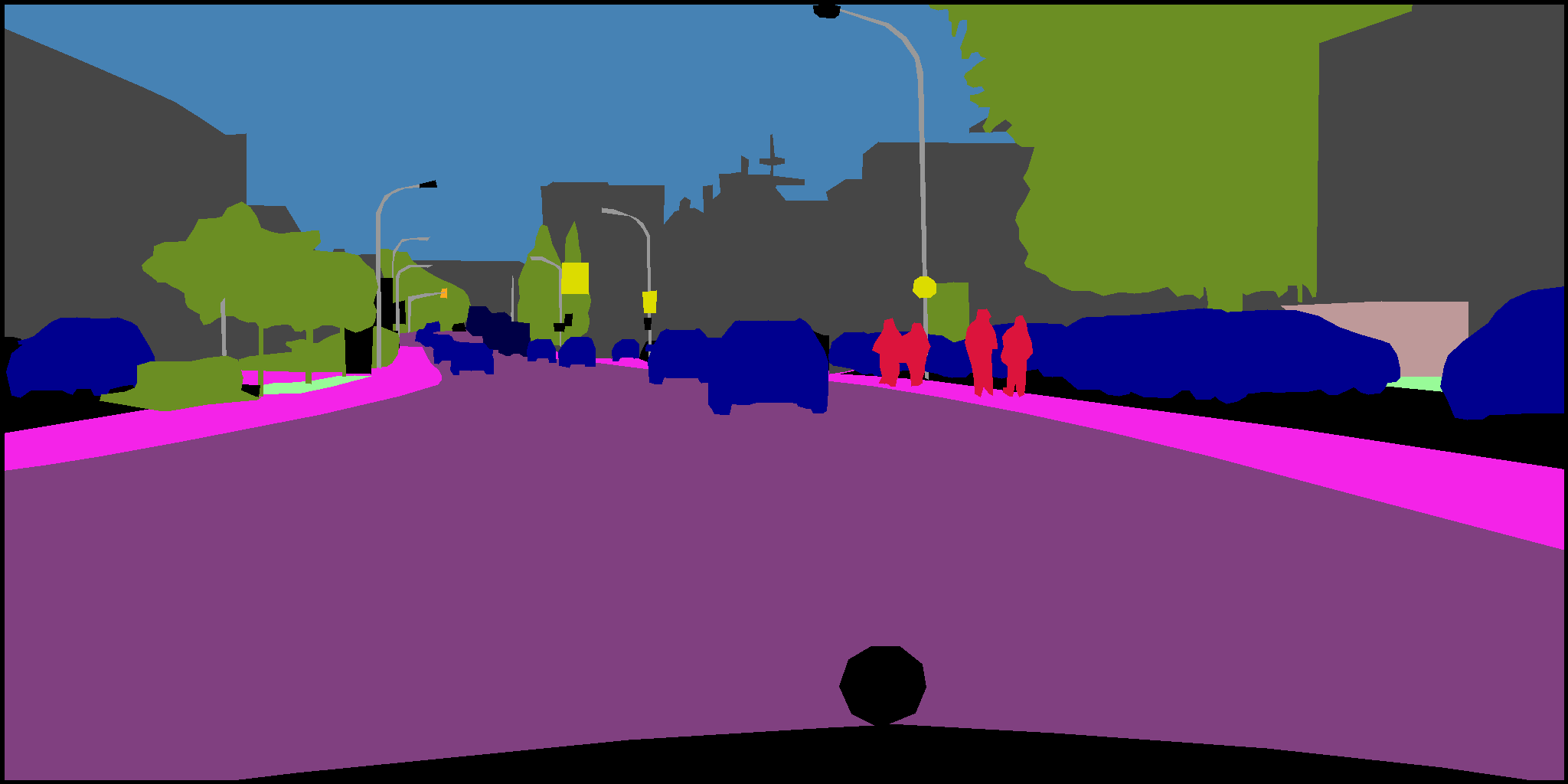}\vspace{8pt}
				\includegraphics[width=\linewidth]{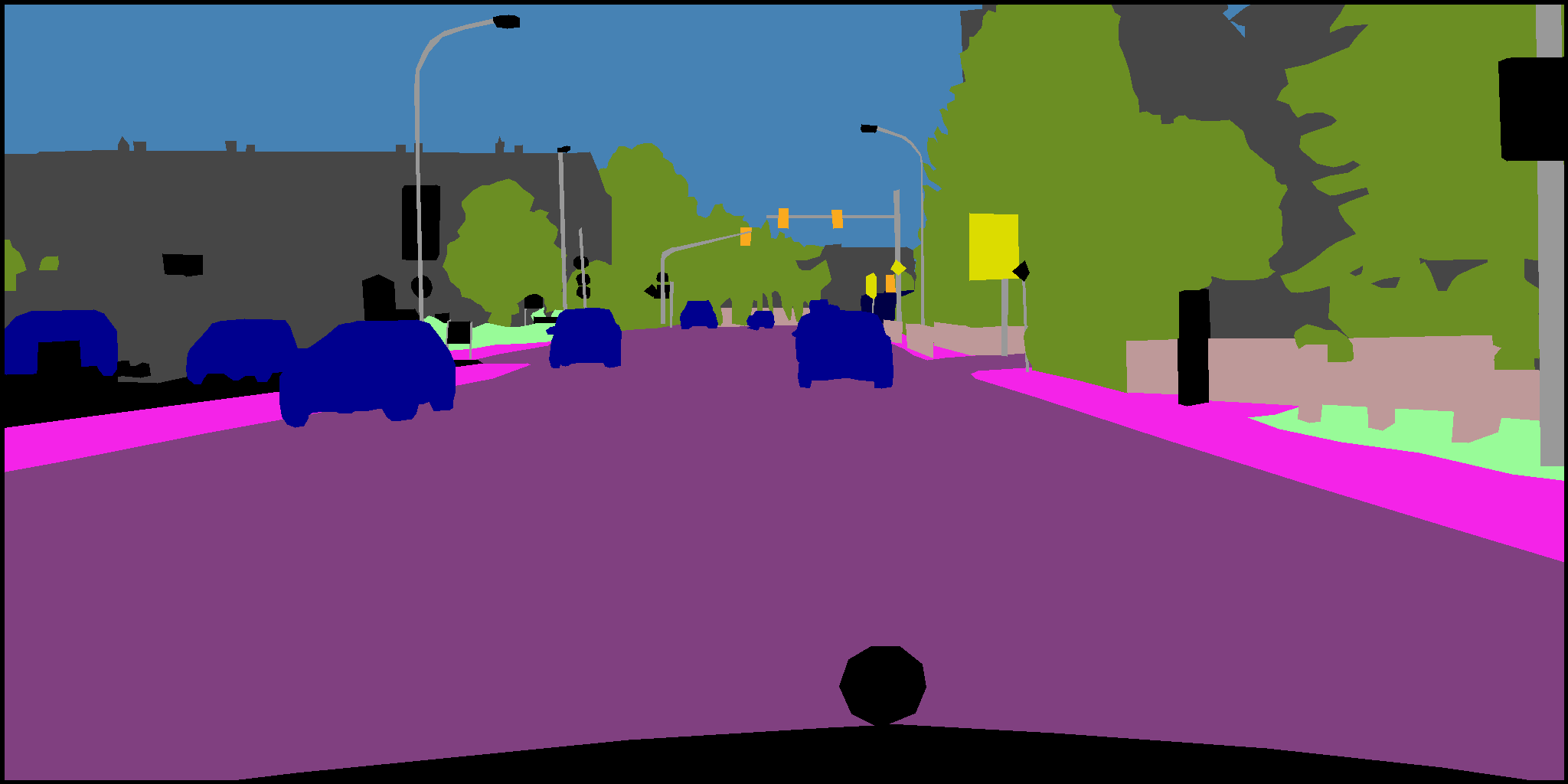}\vspace{8pt}
				\includegraphics[width=\linewidth]{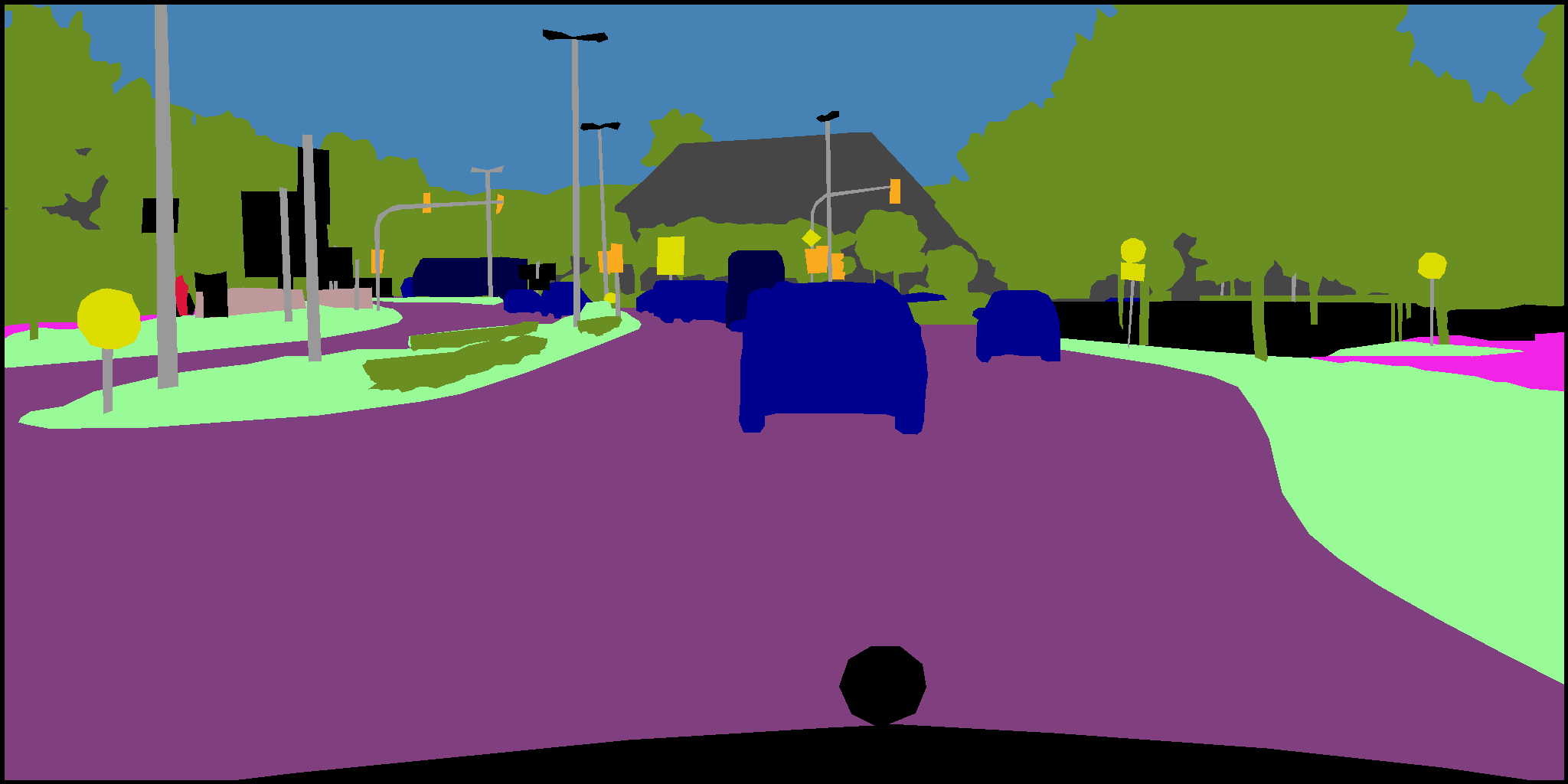}
			\end{minipage}
			\caption{Ground truth}
		}
	\end{subfigure} 	
	\begin{subfigure}[b]{0.3\textwidth}{
			\begin{minipage}[b]{1\linewidth}
				\centering
				\includegraphics[width=\linewidth]{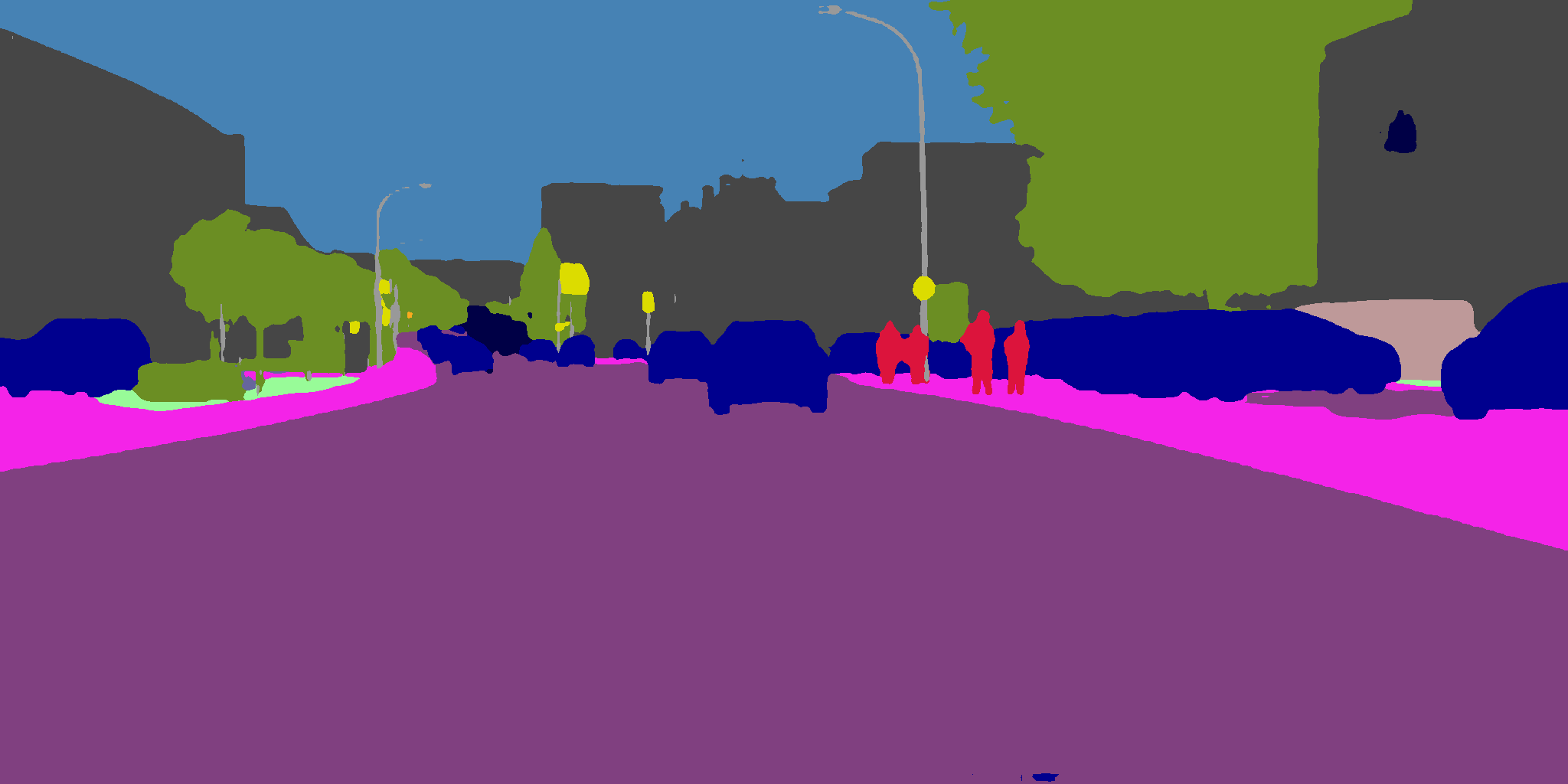}\vspace{8pt}
				\includegraphics[width=\linewidth]{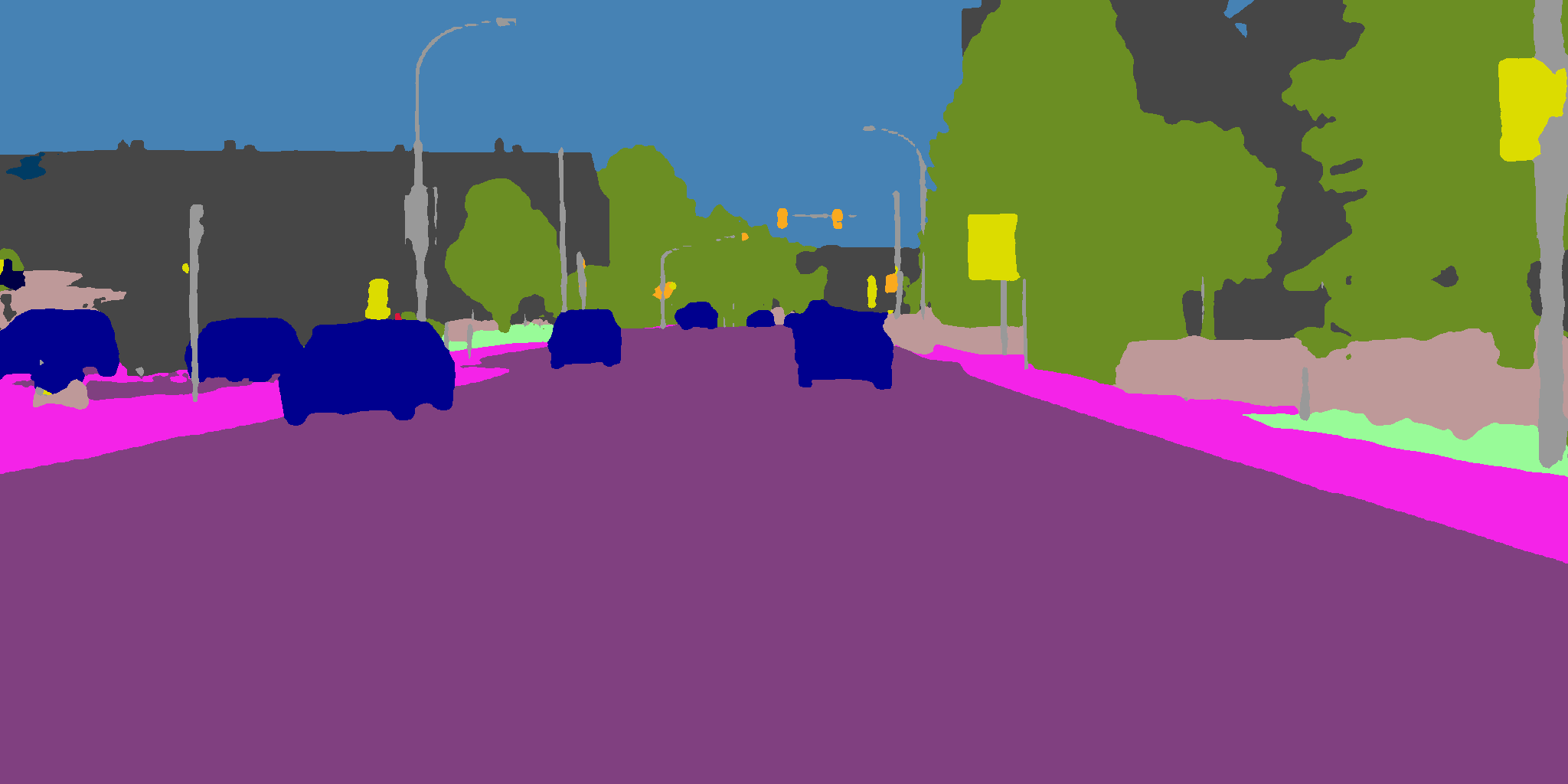}\vspace{8pt}
				\includegraphics[width=\linewidth]{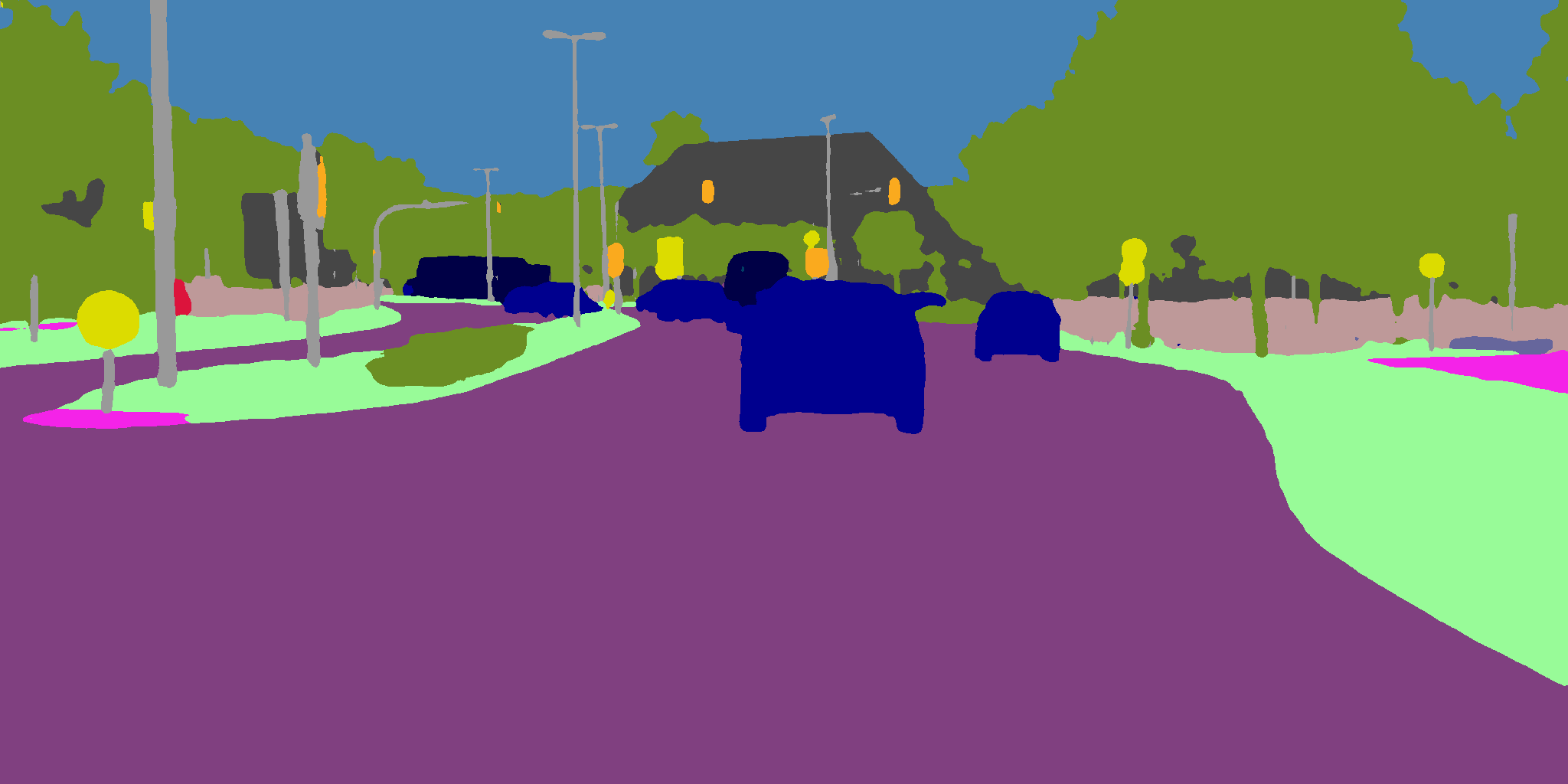}
			\end{minipage}
			\caption{Output image}
		}
	\end{subfigure} 
	\hfill
	
	\vfill
	\caption{Qualitative examples of the segmentation on Cityscapes validation set. From left to right: Input image, ground-truth, and prediction of FRFNet.}
\end{figure*}
\subsection*{Comparison with state-of-the-art works}
\begin{table}[ht]
	\caption{ Evaluation results on the Cityscapes test set}
	\setlength{\tabcolsep}{3mm}
	\centering
	\begin{tabular}{lccccc}
		\hline
		Model & Input Size & FLOPs & FPS & GPU & mIoU(\%)\\
		\hline
		SegNet\cite{2017SegNet} & 640×360 & 286G & 16.7 & TitanX M & 57.0\\
		ENet\cite{paszke2016enet} &640×360&3.8G&135.4&TitanX M&57.0\\
		SQ\cite{treml2016speeding}&1024×2048&270G&16.7&TitanX&59.8\\
		ICNet \cite{zhao2018icnet}&1024×2048&28.3G&30.3&TitanX M&69.0\\
		ERFNet \cite{romera2017erfnet}&512×1024&21.0G&41.7&TitanX M&68.0\\
		\hline
		LedNet\cite {wang2019lednet}&1024x2048&45.84G&24.72&2080Ti&71.3\\
		CABiNet\cite {saksena2020cabinet}&1024x2048&12.0G&76.5&2080Ti&75.9\\
		ShelfNet \cite {zhuang2019shelfnet}&1024x2048&93.69G&44.37&2080Ti&74.8\\
		MSFNet\cite {si2019real}&512x1024&24.2G&119&2080Ti&71.3\\
		\hline
		DABNet \cite{li2019dabnet}&1024×2048&10.46G&58.04&1080Ti&70.1\\
		BiSeNet \cite {yu2018bisenet}&1563x768&103.72G&105.8&1080Ti&71.5\\
		BiSeNetV2 \cite {yu2020bisenet}&512×1024&21.1G&156.0&1080Ti & 72.6\\
		\hline
		FRFNet-slim&512x1024&11.38G&144.4&1080Ti&72.0\\
		FRFNet&512x1024&16.01G&93.8&1080Ti&75.1\\
		\hline
	\end{tabular}
\end{table}

\noindent\textbf{Results on Cityscpaes.}
Table 3 provides a detailed comparison between our method and other architectures based upon the input size, FLOPs count, execution speed (RTX 1080Ti), GPU type and the overall mIOU score on the Cityscapes test sets.
Some visual results of the proposed posed FRFNet are shown in Figure 5. It can be seen that we can achieve high-performance semantic segmentation on Cityscapes. 

\noindent\textbf{Results on Camvid.}
\begin{table}[!h]
	\caption{A Evaluation results on the CamVid test set}
	\setlength{\tabcolsep}{2.5mm}
	\label{tab:2}       
	\centering
	\begin{tabular}{lcc}
		\hline
		Model  & Parameters (M)  & mIoU(\%)\\
		\hline
		ENet \cite{paszke2016enet}&0.36&51.3\\
		SegNet \cite{2017SegNet}&29.5&67.1\\
		FCN-8s \cite{2015Fully}&134.5&65.6\\
		Dilation8\cite{yu2015multi}&140.8&64.7\\
		BiSeNet\cite{yu2018bisenet}&5.8&59.3\\
		ESPNet v2\cite{mehta2019espnetv2}&0.37&59.3\\
		\hline
		Ours&4.02&\textbf{68.2}\\
		\hline
	\end{tabular}
\end{table}
We also evaluate our network on CamVid dataset. As shown in Table 4, our model achieves outstanding performance with small capacity.

\subsection{Comparison with CamVid dataset}
We also evaluate our network on CamVid dataset. As shown in Table 3, our model achieves outstanding performance with small capacity, and it can process a 360×480 CamVid image at the speed of 225 FPS.

\section{Conclusion}
In this article, we describe a FRFNet model to solve the real-time semantic segmentation problem. On the one hand, we designed the backbone network to privilege features. On the other hand, we designed three fusion methods to fully integrate information. The experimental results show that our FRFNet achieves the best compromise on the CityScapes dataset regarding segmentation accuracy and implementation efficiency. Future work includes improving the reception range without adding other calculations, further reducing network weight while maintaining segmentation accuracy.

\section*{Acknowledgements}This research was funded by [the National Natural Science Foundation of China] grant number [No.U1603115], [the National Key Research and Development Program of China] grant number [No.2017-\\YFBO504203], [the Science and Technology Planning Project of Sichuan Province] grant number [No.18SX-\\HZ0054] and [National Engineering Laboratory for Public Safety Risk Perception and Control by Big Data] grant number [PSRPC:No.XJ201810101].
\bibliographystyle{unsrt}  


\end{document}